%% file: main.tex
\documentclass[conference]{IEEEtran}
\IEEEoverridecommandlockouts

\usepackage{cite}
\usepackage{amsmath,amssymb,amsfonts}
\usepackage{balance}
\usepackage{graphicx}
\usepackage{textcomp}
\usepackage{xcolor}
\usepackage{algorithm}
\usepackage{multirow}
\usepackage{diagbox}
\usepackage{threeparttable} 
\def\BibTeX{{\rm B\kern-.05em{\sc i\kern-.025em b}\kern-.08em
    T\kern-.1667em\lower.7ex\hbox{E}\kern-.125emX}}
\begin{document}

\title{Hierarchical Pose Estimation and Mapping\\ with Multi-scale
Neural Feature Fields

}



\author{
    \IEEEauthorblockN{1\textsuperscript{st} Evgenii Kruzhkov\hspace{5.5cm}2\textsuperscript{nd} Alena Savinykh}
    \IEEEauthorblockA{
        \textit{Autonomous Intelligent Systems}\hspace{4.3cm}\textit{Autonomous Intelligent Systems}\\
        \textit{Computer Science Institute VI, University of Bonn\hspace{1.59cm}Computer Science Institute VI, University of Bonn} \\
        Bonn, Germany, ekruzhkov@ais.uni-bonn.de\hspace{2.5cm}Bonn, Germany, asavinykh@ais.uni-bonn.de
    }
    \\
    \IEEEauthorblockN{3\textsuperscript{rd} Sven Behnke}
    \IEEEauthorblockA{
        \textit{Autonomous Intelligent Systems, Computer Science Institute VI – Intelligent Systems and Robotics} \\
        \textit{Center for Robotics, and Lamarr Institute for Machine Learning and Artificial Intelligence, University of Bonn} \\
        Bonn, Germany, behnke@cs.uni-bonn.de
    }
}

\maketitle

\input{chapters/0_abstract}


\input{chapters/1_intro}
\input{chapters/2_related_works}

\input{chapters/3_method}
\input{chapters/4_experiments}

\input{chapters/5_conclusion}

\bibliographystyle{IEEEtran}
\bibliography{main.bib}

\end{document}

%% file: chapters/0_abstract.tex
\begin{abstract}
Robotic applications require a comprehensive understanding of the scene. 
In recent years, neural fields-based approaches that parameterize the entire environment have become popular. 
These approaches are promising due to their continuous nature and their ability to learn scene priors. 
However, the use of neural fields in robotics becomes challenging when dealing with unknown sensor poses and sequential measurements. 
This paper focuses on the problem of sensor pose estimation for large-scale neural implicit SLAM.
We investigate implicit mapping from a probabilistic perspective and propose hierarchical pose estimation with a corresponding neural network architecture. Our method is well-suited for large-scale implicit map representations.
The proposed approach operates on consecutive outdoor LiDAR scans and achieves accurate pose estimation, while maintaining stable mapping quality for both short and long trajectories.
We built our method on a structured and sparse implicit representation suitable for large-scale reconstruction and evaluated it using the KITTI and MaiCity datasets. Our approach outperforms the baseline in terms of mapping with unknown poses and achieves state-of-the-art localization accuracy.

\end{abstract}

%% file: chapters/1_intro.tex
\section{Introduction}

Pose estimation and mapping are essential tasks that have a significant impact on mobile robotics, self-driving cars, virtual reality, and computer vision. These two tasks are closely related, as the quality of one influences the quality of the other. The combined task is commonly known as simultaneous localization and mapping (SLAM). In recent years, researchers have explored novel approaches to high-quality and efficient mapping with accurate pose estimation~\cite{vizzo2023kiss, zhang2014loam, droeschel2018efficient, quenzel2021real}.

Although significant progress has been made in SLAM, it remains necessary to develop novel methods to meet the increasing demands for quality and generalization ability.
For instance, sparse methods~\cite{ferrera2021ov} produce maps that have limited applicability (mainly further localization) because they do not represent dense surfaces or volumes.
On the other hand, dense~\cite{jatavallabhula2020slam} and semi-dense~\cite{engel2014lsd} maps can capture a greater amount of information and finer details allowing numerous downstream tasks. However, they require more computational resources and still have limited expression capability because of their discrete nature. 

Today, numerous robotic applications~\cite{habitatchallenge2023} demand a more advanced understanding of the environment, which calls for the development of novel comprehensive maps.
To address this, some works~\cite{ nieuwenhuisen2010improving, sucar2020nodeslam, wang2021dsp} introduce parameterized elements to the scenes (doors, planes, small items, etc.).
However, these approaches can parameterize only a portion of the scene's objects, which can be sufficient for some scenarios but does not provide a general solution.

\begin{figure}[t!]
\centering
\includegraphics[scale=0.265]{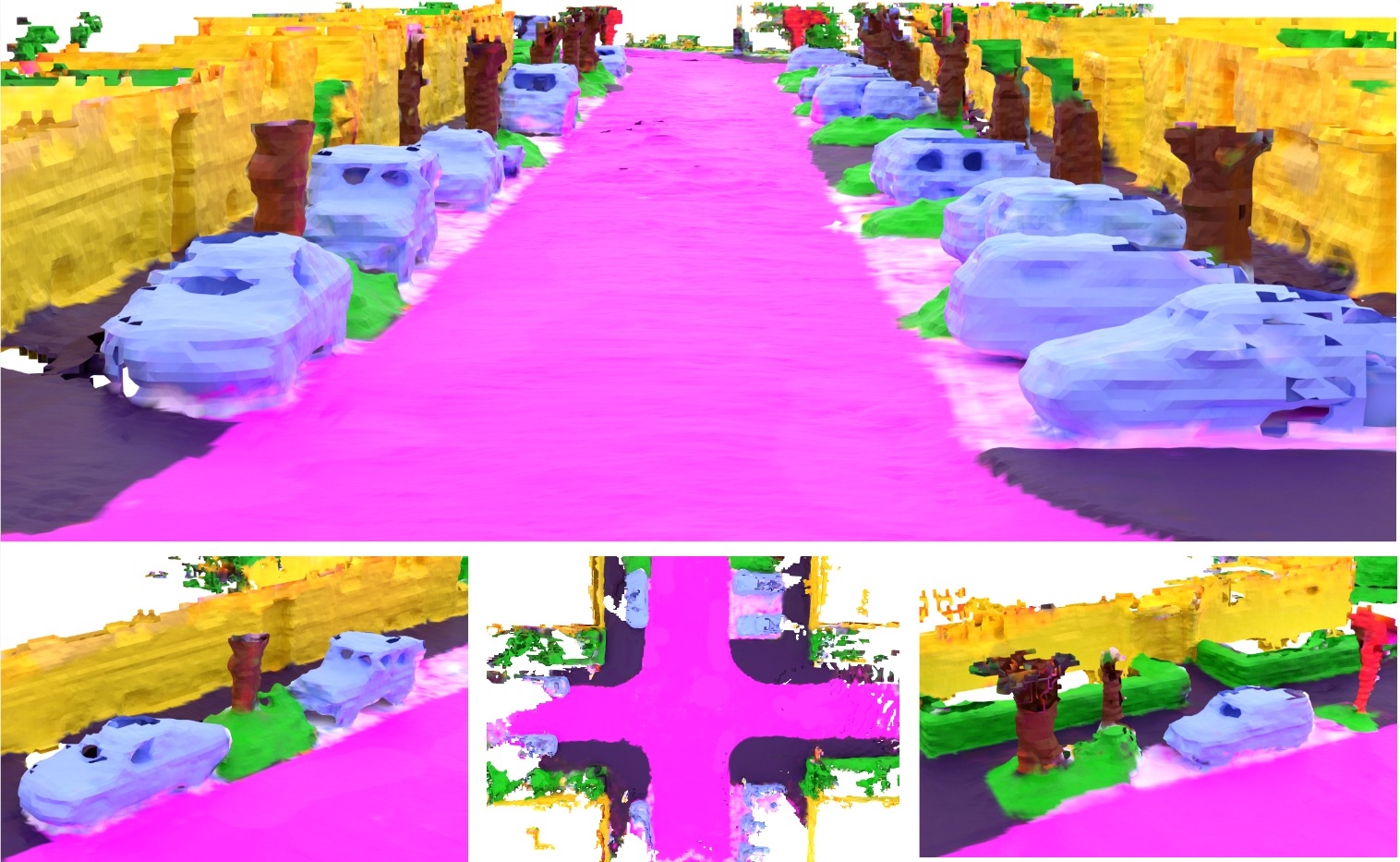}
\caption{
Demonstration of the flexibility of the proposed implicit mapping extended to semantic domain.
We create the map using sequential LiDAR measurements and corresponding semantic labels from the SemanticKITTI dataset~\cite{behley2021ijrr}. Our approach effectively learns a 3D semantic representation from the data, demonstrating its generalization abilities. 
We employ the Marching Cubes algorithm~\cite{lorensen1998marching} to visualize the learned information.}
\label{Fig:semantic-fig}
\end{figure}
 
Recently,  Neural Radiance Fields (NeRF)~\cite{mildenhall2021nerf, adamkiewicz2022vision, kurenkov2022nfomp} have gained attention in the scientific community. 
Although most of these approaches rely on ground truth poses that are unavailable for real-time robotics applications, they still possess useful properties.
For instance, neural fields can parameterize the entire scene from plain geometry to fine-grained details~\cite{tancik2022block, muller2022instant, lin2021barf, rosu2023permutosdf}.
Moreover, this representation can easily be extended to additional levels of scene understanding~\cite{Zhi:etal:arxiv2021}.
Figure~\ref{Fig:semantic-fig} illustrates the semantic mapping achieved by the proposed approach that is augmented to represent semantic information alongside 3D geometry (Section~\ref{network-architecture}).


In this paper, we advance the current state of research on neural implicit representations for robotic applications.
Specifically, we focus on implicit representations of large-scale environments and simultaneous pose estimation inside neural fields.
We provide a probabilistic interpretation of implicit mapping and introduce coarse-to-fine pose optimization based on sequential LiDAR data for large-scale neural field maps, enabling operations in street-sized environments.  

\noindent In summary, we present in this work:
\begin{itemize}
    \item structured sparse implicit mapping and pose estimation with probabilistic interpretation,
    \item hierarchical pose optimization based on the octree map structure, and
    \item evaluation of our proposed approach relative to state-of-the-art methods.
\end{itemize}

%% file: chapters/2_related_works.tex
\section{Related Works}
CodeSLAM~\cite{bloesch2018codeslam} is one of the pioneering works in SLAM that proposes a compact auto-encoder based implicit environment representation. However, in recent years, NeRF-based architectures~\cite{mildenhall2021nerf} have gained popularity in the field of implicit mapping, leading to the development of novel SLAM algorithms. Deng \textit{et al.}~\cite{deng2022depth} demonstrated that providing additional depth information improves the training convergence of such architectures. For instance, iMap~\cite{sucar2021imap} is one of the first approaches that uses neural radiance fields and an RGB-D sensor for SLAM, but it can only operate in small environments.

Several studies have advanced neural fields to encompass larger scales. 
For instance, KiloNeRF~\cite{reiser2021kilonerf} proposes the use of tiny voxel-assigned neural networks to accelerate learning and inference. Similarly, MeSLAM~\cite{kruzhkov2022meslam} utilizes coupled neural field blocks to achieve scalability.
Müller \textit{et al.}~\cite{muller2022instant} introduce multiresolution hash encoding, which considerably improved training speed and network generalization capabilities, but it is not easily applicable to unbounded scenes. In contrast, Nice-SLAM~\cite{zhu2022nice} implements a grid structure that enables mid-size environment mapping, but it is inefficient for open space scenes.

Studies including \cite{zhu2022nice, sucar2021imap, kruzhkov2022meslam} sample points along the rays and aggregate predictions at these sampled points to derive a loss for training neural fields. However, this approach can be computationally inefficient for open-space environments. 
In contrast, we consider a probabilistic interpretation of the mapping task (Sec.~\ref{method-mapping}) and derive direct optimization to avoid computationally expensive volume rendering.
Our method utilizes octree-based feature grids~\cite{takikawa2021neural}, making it suitable for open-space environments.

Active research on gradient-based pose optimization is conducted in works~\cite{teed2021tangent, wang2021nerf}. BARF~\cite{lin2021barf} proposes coarse-to-fine pose optimization, which facilitates neural field training with imprecise camera poses. However, this method assumes the instant availability of all data and is designed for specific positional encoding~\cite{tancik2020fourfeat}. Pure localization in neural fields is also studied, as seen in \cite{wiesmann2023locndf, kuang2023ral}, which focus on Monte Carlo localization~\cite{fox1999monte} in implicit maps.

Large-scale neural rendering reconstruction is addressed in \cite{tancik2022block, martin2021nerf}, but these approaches are not easily adaptable to SLAM. Moreover, approaches based on \cite{mildenhall2021nerf} suffer from artifacts in the reconstructed 3D geometry. To address this issue, Azinovi{\'c} et al.~\cite{azinovic2022neural} propose to directly predict a truncated signed distance function (TSDF). Zhong et al.~\cite{zhong2023shine} combine an efficient octree structure with the prediction of TSDF to achieve a precise large-scale 3D reconstruction. 

The work most similar to ours is SHINE-Mapping~\cite{zhong2023shine}, which introduces a mapping-with-known-poses method.
We use it as a baseline to estimate the mapping quality of our approach.
In contrast to related works, our method enables large-scale implicit mapping and hierarchical pose estimation in neural fields. It handles sequential 3D LiDAR data and does not require ground-truth poses, making it suitable for practical robotic applications.

%% file: chapters/3_method.tex
\begin{figure*}[ht]
\centering
\includegraphics[scale=0.241]{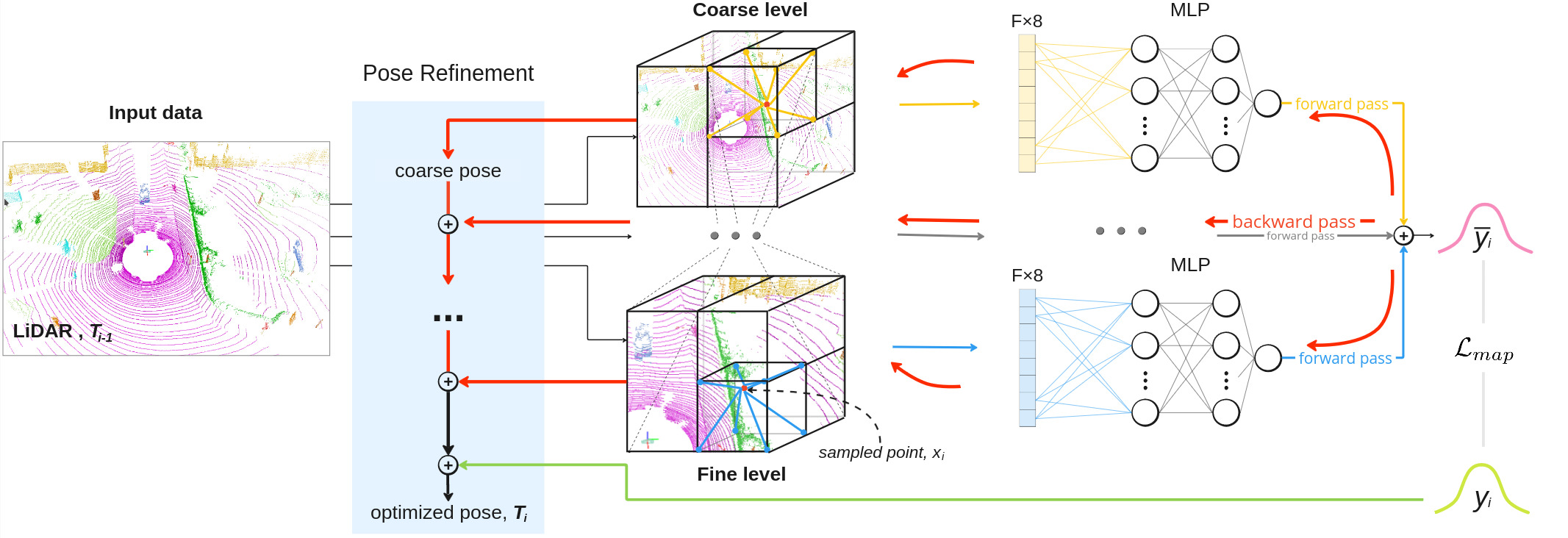}
\caption{Overview of the proposed approach. Our method utilizes neural fields consisting of learnable $F$-dimensional features stored in the corners of an octree-based structure. Each trainable octree level is associated with tiny MLP networks. During the forward pass, the LiDAR measurements are transformed to world coordinates using an initial transformation $T_{i-1}$. 
The features of the voxel corners are then weighted based on the relative position of the sampled point $x_i$ (red) from the measurements.
These weighted features are concatenated and fed to the corresponding MLP. The predictions of all layers are accumulated to generate the final occupancy probability $\bar{y}_i$ for the sampled points. During the backward pass, the gradient values are backpropagated to optimize the transformation $T_{i-1}$ toward $T_i$ in two ways: directly through measurements $y_i$ and hierarchically though $\bar{y}_i$ (Sec.~\ref{pose_optimization}).} 
\label{Fig:pipeline}
\end{figure*}

\section{Method}

This section describes the components of our method.
We first introduce a probabilistic interpretation of the proposed approach.
Then we describe the proposed neural network architecture in depth.  Finally, we present the details of our pose estimation in a hierarchical, coarse-to-fine manner.  

Fig.~\ref{Fig:pipeline} illustrates the core components of our method. 
The proposed network architecture is trained to represent the map with varying levels of detail using 3D LiDAR data as input.

Based on the previous pose estimations, we optimize the current sensor pose through gradient flow defined by loss between current LiDAR measurements and the learned implicit map.
As a result, our approach performs simultaneous pose estimation and implicit mapping based on sequential LiDAR data.

\subsection{Mapping}\label{method-mapping}
Information about the environment is obtained from measurements of noisy sensors. We consider the environment to be a probability density function (PDF) of the measured noisy values. We employ a neural field-based model to approximate the PDF of the environment. To learn the model, we minimize the KL divergence between the approximated and real PDFs, denoted $q(x, w)$ and $p(x)$, respectively. 
Minimizing $D_{KL}$ is equivalent to minimizing cross-entropy~\cite{zhang2018generalized} and maximizing the log-likelihood of $q(x,w)$:

\begin{equation}\label{kl-divergened}
\min_{w} D_{KL}\left(\frac{p(x)}{q(x, w)}\right) \sim \max_{w} \frac{1}{N}\sum_{i=1}^{N}log(q(x_i, w)),
\end{equation}

where $w$ represents the weights of the neural field model, $x$ is a 3D spatial coordinate and $N$ is the number of sampled points.   

We assume that the sensor measurements $y(x_i)$ have normally distributed noise with mean $\mu$ and variance $\sigma^2$.
Then the measured environment has normally distributed 
PDF:
\begin{equation}\label{eq-normal-noise}
q(x_i,w| \mu,\sigma) = \frac{1}{\sigma\sqrt{2\pi}}e^{-\frac{(y(x_i)-\mu)^2}{2\sigma^2}}.
\end{equation}

Substituting Eq.~\ref{eq-normal-noise} into Eq.~\ref{kl-divergened}, the likelihood maximization becomes a root mean square error minimization (RMSE) problem:
\begin{equation}\label{Loss-mapping}
\mathcal{L}_{map} = \frac{1}{N}\sum_{i=1}^{N}(y(x_i) - \bar{y}(x_i))^2,
\end{equation}
where $\bar{y}(x_i)$ represents the prediction of the model at coordinate $x_i$ and $y(x_i)$ represents the mean $\mu$ of the measurements. 
We choose the signed distance function (SDF) as the environment representation.
We acquire environment observations $y(x_i)$ by sampling points around LiDAR point clouds and approximating $y(x_i)$ as the distance to the closest point in the LiDAR measurements.
Both $y(x_i)$ and $\bar{y}(x_i)$ are direct values of $x_i$ obtained without volume rendering.

\input{tables/localization}
\subsection{Network Architecture}\label{network-architecture}
We employ learnable $F$-dimensional features that are fed to MLPs as our neural field representation. 
These features are stored in corners of an octree structure (Fig.~\ref{Fig:pipeline}).
Although the final reconstruction is a dense SDF, the octree sparsely stores learnable features only for the voxels where LiDAR measurements were observed.
We use different octree levels and explicitly train them to learn coarse, mid-level and fine representations of the environment.

First, each layer is independent and has its own MLP neural network.
The level's learnable features, combined with spatial coordinates of a queried point (Sec.~\ref{pose_optimization}), serve as input to the network that infers the occupancy probability for the point.
The combined features are concatenated, which, based on our observations, yields better prediction quality compared to the mean and sum input reductions.

Second, each MLP is explicitly guided to represent only its level details during mapping.
To achieve this, we deactivate all layers except the coarsest one for the first iterations of Eq.~\ref{Loss-mapping} optimization. 
Then, we activate finer levels and their corresponding MLPs one by one, with an equal iteration interval.
The predictions from the newly activated levels are added to the coarser ones, guiding the learning of the finer details. 
We follow this technique for each mapping step.

\begin{figure}[t!]
\centering
\includegraphics[scale=0.245]{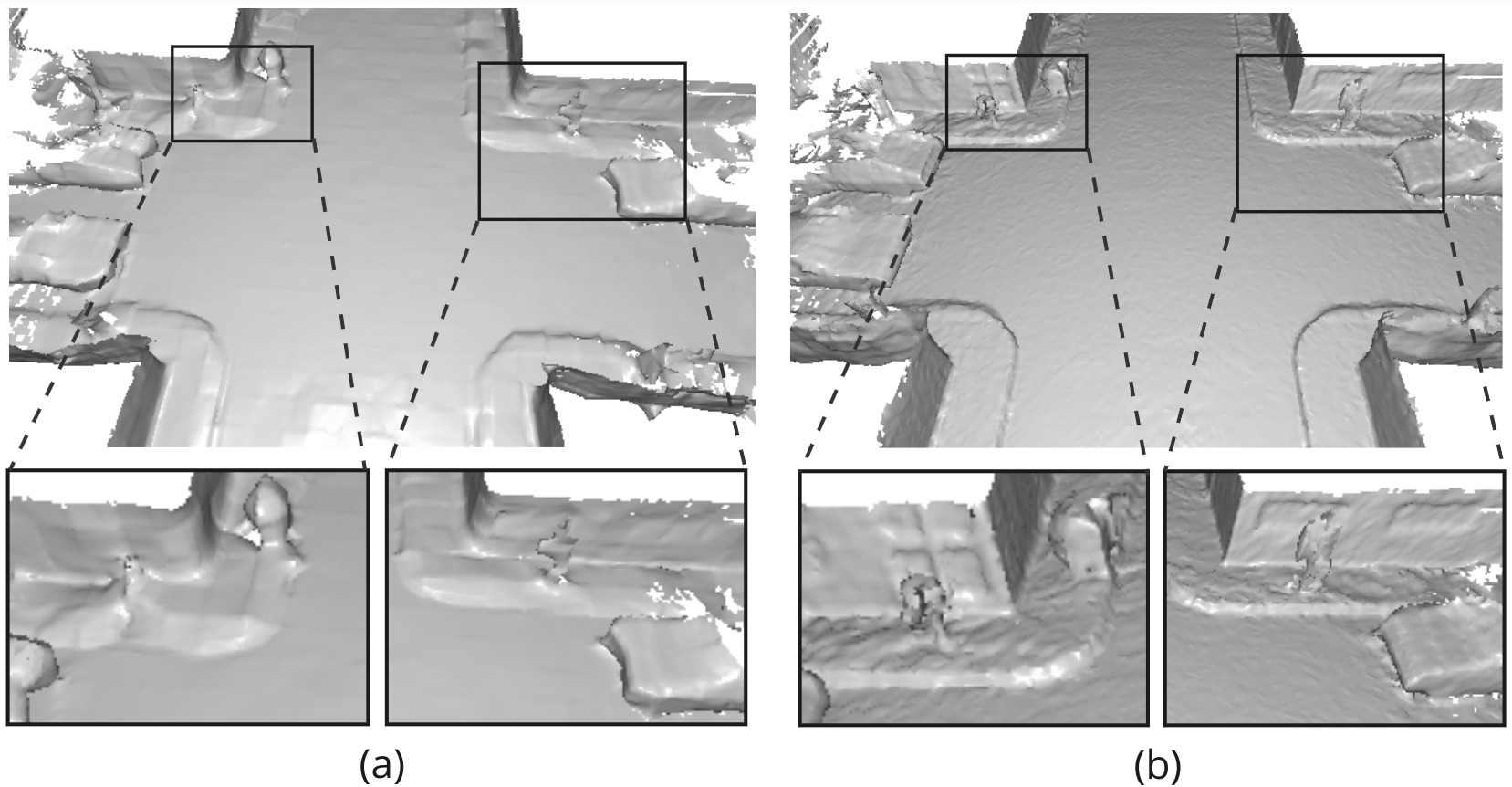}
\caption{The coarse and final representations learned by our proposed approach, reconstructed using the Marching Cubes algorithm. (a) shows the representation learned solely by the coarse level MLP, while (b) displays the final representation that includes high-frequency features from the finer levels.}
\label{Fig:hierarchical-predictions}
\end{figure}

Finally, we sum the predictions from all levels to obtain both low- and high-frequency details on the final map. 
Fig.~\ref{Fig:hierarchical-predictions} demonstrates the difference between the coarse-level prediction and the final prediction. The coarse level represents rough geometry, while the finer levels capture high-frequency details, resulting in improved map quality when combined together. 
With the proposed technique, predictions of the finer levels are guided to complement the coarser ones without redundantly learning the geometry representations of the previous levels.

We observed that finer octree levels are resistant to catastrophic forgetting if the input measurements are dense and the close measurements have sufficient density.
Moreover, the close measurements overlap with the environment which is already mapped from the past observations. 
Therefore, we address catastrophic forgetting differently for close and far LiDAR measurements.
For the far points, we tackle the problem regularizing the learnable levels' features similar to Zhong et al.~\cite{zhong2023shine}:
\begin{equation}
\mathcal{L}_{k, reg} =  \gamma_{k} \sum \Omega (\textbf{w}^{t} - \textbf{w}^{t-1}),
\end{equation}
where $\textbf{w}^{t}$ and $\textbf{w}^{t-1}$ represent the current and previously converged  levels' features, respectively, $\Omega$ are importance weights proposed by \cite{zhong2023shine}, and $\gamma_{k}$ is a level-$k$ regularization weight introduced in our work.
We propose higher weights $\gamma_{k}$ for coarse levels to prevent forgetting of the whole geometry and smaller weights for the finer levels to allow updating of details.  

For close measurements, we regularize outer octree voxels by sparsely reconstructing the points sampled at the fine level and then adding them to the measurements as replay data, thus employing the hierarchical structure of our implicit representation.
Our final loss for joint training of the levels' features and MLPs becomes:

\begin{equation}
\mathcal{L} = \mathcal{L}_{map}  + \sum_{k} \mathcal{L}_{k, reg}.
\end{equation}

We observe that due to the sparse nature of the octree architecture, we can extend it to additional domains without drastically increasing the consumed memory.
To generate Fig.~\ref{Fig:semantic-fig}, we extended the dimensions of the input with encoded semantic labels, increased the size of the features stored in the corners of the octree, and optimized the model to reconstruct encoded semantic labels together with the geometry using the same Eq.~\ref{Loss-mapping}.

\subsection{Pose Estimation}\label{pose_optimization}

We perform simultaneous localization by optimizing the sensor pose in the neural fields.
Having approximation $q(x, w)$ of the environment and current measurements $y(x_i)$, we optimize the sensor pose $T_i$ which is used to project measurements from the sensor's local coordinates to the global coordinates.
According to Eq.~\ref{Loss-mapping}, it can be done through both  $y(x_i)$ and  $\bar{y}(x_i)$ but the path to $T_i$ through the measurements $y(x_i)$ is more straightforward than through model prediction $\bar{y}(x_i)$.
The meaning of such optimization is to find the position of the sensor where its observations correspond to the mapped environment $q(x, w)$ with the smallest RMSE.
We use the previously estimated pose $T_{i-1}$ as the initial one for each new optimization step.



During the forward pass, we weigh the features of each octree level according to the relative position of the sampled point $x$ within its voxel before feeding them to the network:
\begin{equation}
z_{j} = h_{j}\cdot \prod (1 - x_s - \lfloor x_s \rfloor),
\end{equation}
where $z_{j}$ represents the weighted $j$-corner features, $h_{j}$ are the corner's $F$-dimensional features, and $x_s$ is the sampled point position scaled to the level's grid resolution.

The weighted features of the same voxel are concatenated and fed into the corresponding MLP model, as shown in Fig.~\ref{Fig:pipeline}.
With this structure, we propagate the gradient to the sensor pose directly through $y(x_i)$.


We additionally employ the proposed hierarchical architecture (Sec.~\ref{network-architecture}) and perform coarse-to-fine pose estimation.
We first perform a coarse pose estimation through the coarse level and then gradually activate the finer levels to refine the pose through more detailed maps.
According to our observations, pose optimization is more stable when performed in such a coarse-to-fine manner.  
Our final pose estimation consists of the direct (Fig.~\ref{Fig:pipeline}: green line) and coarse-to-fine optimizations (Fig.~\ref{Fig:pipeline}: red lines) performed concurrently.

%% file: tables/localization.tex
\begin{table*}[ht]
\caption{Localization quality estimation on KITTI dataset}
\label{table_loc}
\begin{center}
\begin{tabular}{c||cc || cc || cc ||ccc}
\hline
\multirow{2}{*}{\backslashbox{Method}{KITTI}} &
\multicolumn{2} {c||}{Seq. 00 (3711 m)}  & 
\multicolumn{2}{c||}{Seq. 05 (2202 m)} &
\multicolumn{2}{c||}{Seq. 03 (567 m)} &
\multicolumn{2}{c}{Seq. 07 (693 m)} \\
& 
\textbf{ATE [m]} &  \textbf{ATE [\%]}  &
\textbf{ATE [m]} &   \textbf{ATE [\%]}  &
\textbf{ATE [m]} &   \textbf{ATE [\%]} &
\textbf{ATE [m]} &   \textbf{ATE [\%]}   \\
\hline
ICP & 
37.55 & 1.04 & 
11.80 & 0.53 & 
5.06 & 0.89 & 
7.68 & 1.11   \\
KISS-ICP & 
6.28 & 0.17 & 
\textbf{1.94} & \textbf{0.09} &
3.35 & 0.59 & 
\textbf{0.758} & \textbf{0.11}    \\
\hline
Ours & 
\textbf{4.87} & \textbf{0.13} & 
4.07 & 0.18 &
\textbf{1.47} & \textbf{0.26} & 
0.87 & 0.13  \\
\hline
\end{tabular}
\end{center}
\end{table*}

%% file: chapters/4_experiments.tex
\section{Experiments}

We evaluated the accuracy of the proposed hierarchical pose estimation in the KITTI dataset~\cite{behley2021ijrr}, which contains outdoor LiDAR sequences of different lengths.
To assess the localization accuracy, we use the absolute trajectory error (ATE).
We also measure the influence of pose estimation on mapping quality using the synthetic MaiCity dataset~\cite{vizzo2021icra}.

Throughout the experiments, we used unified parameters for localization and mapping. We set the feature dimension $F=3$ and use $3$ octree levels with learnable features. All tiny MLPs have 2 hidden layers with 32 neurons each. 

Our approach takes $50$ iterations for mapping. For pose estimation, we set $80$ iteration: $20$ iterations for coarse estimation, $20$ iterations with the middle level activated, and the remaining $40$ iterations with all levels activated.

\subsection{Pose Estimation}

\input{tables/mapping}

The aim of this experiment is to verify that the proposed approach can effectively perform continuous localization throughout the whole sequence with decent and stable accuracy.
We compare its performance with point-to-plane ICP~\cite{chen1992object} and state-of-the-art KISS-ICP~\cite{vizzo2023kiss}. The parameters of these methods are fixed for all sequences.

The results are presented in Tab.~\ref{table_loc}.
The proposed approach is superior to ICP and comparable to the state-of-the-art KISS-ICP. Based on the experimental results, our method can provide consistent localization results for both short (Seq. 03 and Seq. 07) and long sequences (Seq. 00 and Seq. 05).

\begin{figure}[t!]
\centering
\includegraphics[scale=0.33]{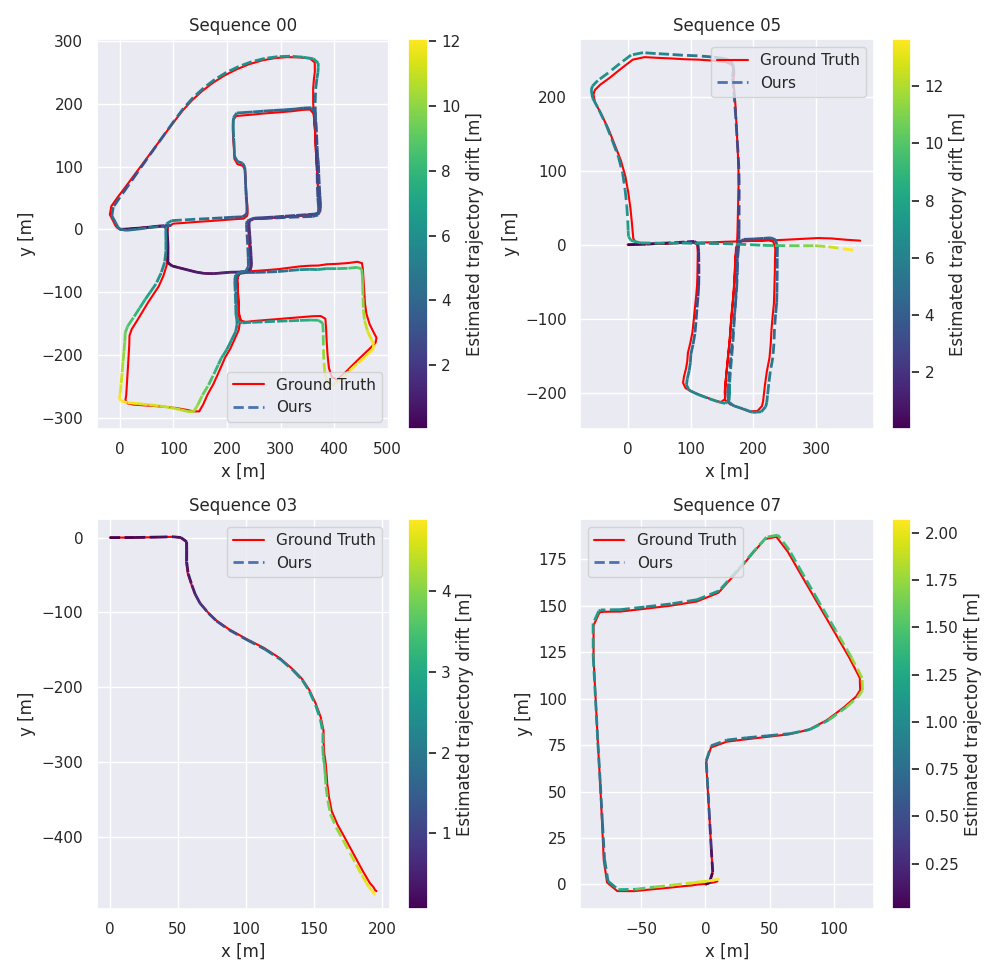}
\caption{Estimated trajectories of the proposed approach on the KITTI dataset. 
The colorbar visualizes the distance between the poses of the estimated and ground truth trajectories.}
\label{Fig:trajectories}
\end{figure}

Fig.~\ref{Fig:trajectories} shows our estimated trajectories and ground truth data for each sequence. We observe that the estimated trajectory usually follows the ground truth precisely. Further analysis reveals that the majority of residuals come from inaccuracy in orientation optimization around the pitch angle.

\begin{figure}[t!]
\centering
\includegraphics[scale=0.315]{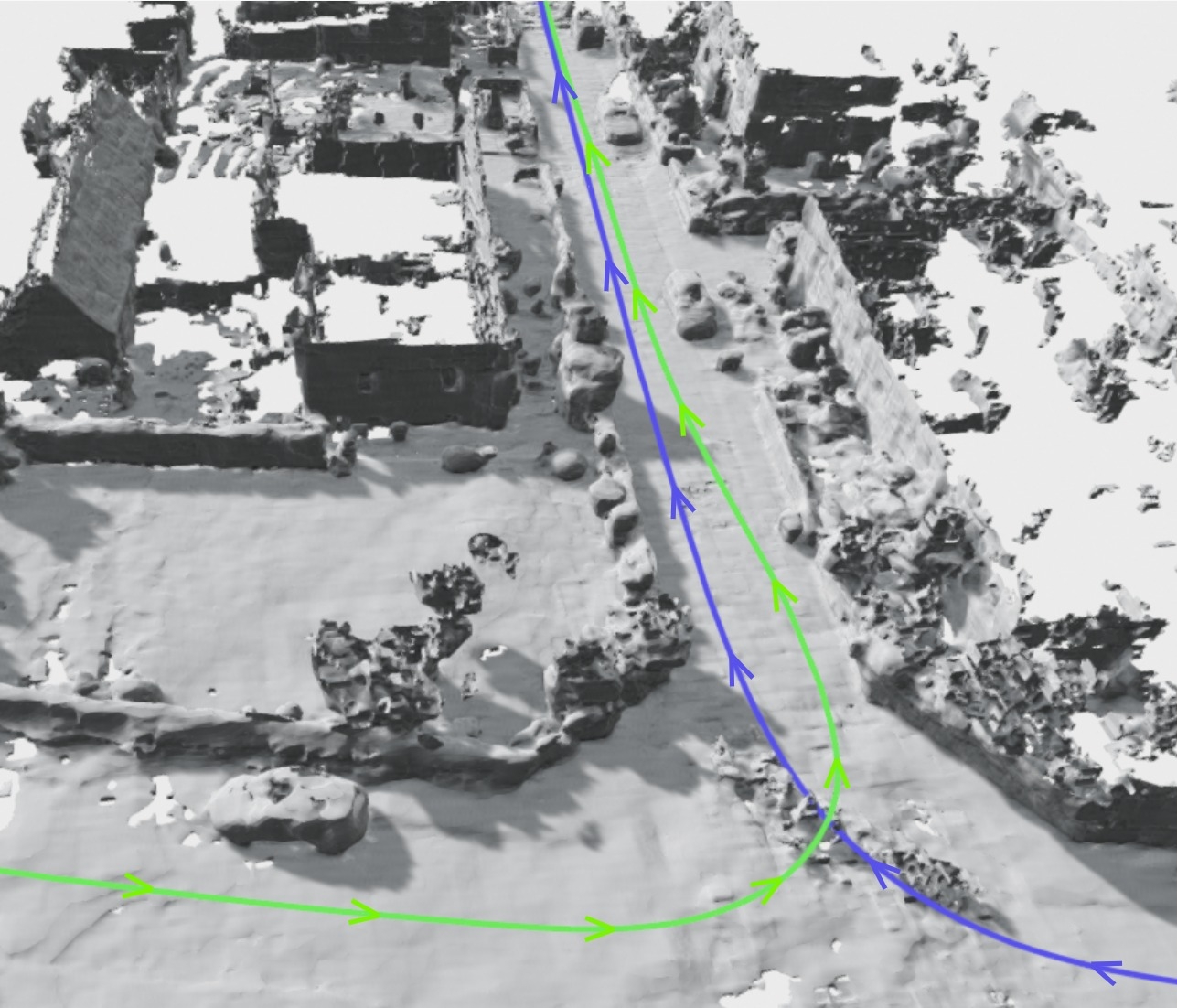}
\caption{Revisiting of previously mapped area. 
The map is initialized during the first visit of the region (green). During the second traversal (blue), the path successfully converges to the path taken during the first visit.
Both traversals have the same motion direction.
}
\label{Fig:loop-vlosure}
\end{figure}

The proposed approach possesses good capabilities to handle places previously visited. For instance, trajectories in Seq.~00 and Seq.~05 shown in Fig.~\ref{Fig:trajectories} have higher accuracy (darker color) when the place is visited more than once. Thus, our method can successfully re-localize itself and decrease localization error when operating in re-visited environments. We illustrate this in Fig.~\ref{Fig:loop-vlosure} which shows a street traversed twice in the same direction. The corresponding trajectories are colored in different colors. The blue trajectory corresponds to the second traversal and it converges to the first visit that is shown in green.

\subsection{Mapping}
This experiment verifies the quality of the map in short and long sequences of the MaiCity dataset~\cite{vizzo2021icra}. We compare the proposed approach against SHINE-Mapping~\cite{zhong2023shine}, which is similar to our mapping. The comparison results in terms of accuracy (Acc.), completeness (Comp.), and Chamfer distances (C-L1, C-L2) are presented in Tab.~\ref{table_mapping}.

We feed the LiDAR data to algorithms in a sequential manner since future measurements are unknown in the case of real-time operation. SHINE-Mapping is a pure mapping approach; thus it uses ground truth poses to infer a map. For comparison, we use SHINE-Mapping with ground truth poses (the best scenario) and poses estimated by the ICP method. 

According to Tab.~\ref{table_mapping}, for the short Seq. 01 our approach achieves more accurate mapping results compared to both variants of SHINE-Mapping. This occurs because SHINE-Mapping is a reconstruction approach that demonstrates the best results when all measurements are available at the same time, while our approach is initially designed to handle sequential data at each time step and benefits from hierarchical optimizations.
As expected, the reconstruction quality of our method is slightly worse than SHINE-Mapping with ground truth poses for the long Seq. 00, but the accuracy values of both methods are still close.
The increased mapping error in the longer sequence is caused by an accumulated localization error that slowly deforms the map. However, the mapping accuracy of our method has similar values in both short and long sequences.
All these highlights the stability of the proposed pose estimation throughout sequences with various length.

Based on the experimental results, both SHINE-Mapping and our method are capable of providing an accurate map. However, SHINE-mapping quality depends on the localization accuracy estimated by third-party algorithms, while the proposed approach has stable built-in localization.

\subsection{Real-Time Operation Discussion}
The efficiency of our approach depends primarily on the size of the environment, the number of optimization steps per mapping and tracking, and the amount of data in the current measurement.
We estimate the time efficiency on a laptop with an embedded NVIDIA GeForce RTX 3070 Ti and 16 GB of RAM memory, noting that modern autonomous robots can have even more powerful computational resources.
For large KITTI sequences (approx. 4000\,m), we use the fine level is 13th level of the octree with a resolution of $0.25$ meters. In such conditions, our approach can operate at a frequency of 3\,fps.
For smaller environments (approx. 50\,m), we achieve the frequency of 5\,fps by using the 10th level as the finest level, with a resolution of 0.05\,m. 

Performance can be further increased by adjusting the parameters for each specific case. Efficient operation was achieved thanks to the sparse octree structure, depth-guided training, and the avoidance of costly volume rendering.

%% file: tables/mapping.tex
\begin{table*}[ht]
\caption{Mapping quality resutls on the MaiCity dataset}
\label{table_mapping}
\begin{center}

\begin{threeparttable}
\begin{tabular}{c||cccc || cccc }
\hline
\multirow{2}{*}{\backslashbox{Method}{MaiCity}} &
\multicolumn{4}{c||}{Seq. 00 (700 m)}  & \multicolumn{4}{c}{Seq. 01 (100 m)} \\
 & 
 \textbf{Acc.} &  \textbf{Comp.} & \textbf{C-L1}\tnote{1}  & \textbf{C-L2}\tnote{2} & 
 \textbf{Acc.} & \textbf{Comp.} &  \textbf{C-L1}\tnote{1} & \textbf{C-L2}\tnote{2} \\
\hline
SHINE (GT) & 
\textbf{0.28} & \textbf{0.12} & \textbf{0.20} & \textbf{0.30} &
0.26 & 0.07 & 0.17 & 0.26  \\
SHINE (ICP) & 
5.92 & 0.46 & 3.19 & 5.90 &
0.35 & 0.20 & 0.28 & 0.36  \\
\hline
Ours & 
1.0 & 0.29 & 0.65 & 1.2 &
\textbf{0.11} & \textbf{0.07} & \textbf{0.09} & \textbf{0.17} \\
\hline
\end{tabular}
 \begin{tablenotes}
   \item [1] Chamfer-L1 distance.
   \item [2] Chamfer-L2 distance.
 \end{tablenotes}
\end{threeparttable}
\end{center}
\end{table*}

%% file: chapters/5_conclusion.tex
\section{Conclusion} 
\label{sec:conclusion}
\balance

In this work, we present a pipeline for simultaneous mapping and pose estimation in structured implicit representations. The proposed neural network architecture, which includes the training of features in the octree corners on different levels and following hierarchical pose optimization, is the core for achieving a large-scale yet detailed map without ground-truth poses. Our approach works with real-world LiDAR data stream data and operates at 3-5 frames per second. 

We perform an evaluation of the proposed approach on public datasets. It shows that our method is capable of simultaneous localization while preserving high mapping quality for large-scale environments. The localization accuracy verified in the KITTI dataset is on par with the state-of-the-art performance demonstrated by KISS-ICP.


To validate mapping quality, we compare our method with a state-of-the-art reconstruction baseline approach on the MaiCity dataset. Experimental results show that our mapping is superior when sequential LiDAR data is utilized.

\section*{Acknowledgment}
This research was funded by the German Federal Ministry of Education and Research (BMBF) in the project WestAI -- AI Service Center West, grant no. 01IS22094A.